# Political representation bias in DBpedia and Wikidata as a challenge for downstream processing

Özgür Karadeniz[1 *], Bettina Berendt[2,3,1 *], Sercan Kıyak[1], Stefan Mertens[1], Leen d'Haenens[1]

[1] KU Leuven, Belgium; [2] TU Berlin, Germany; [3] Weizenbaum Institute, Germany
* corresponding authors {ozgur.karadeniz|bettina.berendt}@kuleuven.be

**Abstract:** Diversity Searcher is a tool originally developed to help analyse diversity in news media texts, relying on a form of automated content analysis and thus rests on prior assumptions and depends on certain design choices related to diversity and fairness. One such design choice is the external knowledge source(s) used. In this article, we discuss implications that these sources can have on the results of content analysis. We compare two data sources that Diversity Searcher has worked with – DBpedia and Wikidata – with respect to their ontological coverage and diversity, and describe implications for the resulting analyses of text corpora. We describe a case study of the relative over- or under-representation of Belgian political parties between 1990 and 2020 in the English-language DBpedia, the Dutch-language DBpedia, and Wikidata, and highlight the many decisions needed with regard to the design of this data analysis and the assumptions behind it, as well as implications from the results. In particular, we came across a staggering over-representation of the political right in the English-language DBpedia.

# 1 Introduction

Originally developed as a tool to help analyse diversity in news media texts, Diversity Searcher is now ready to be used by the general public and in the context of public libraries. The Diversity Searcher, presented in section 2, relies on a form of automated content analysis and thus rests on prior assumptions [1] and depends on certain design choices related to diversity and fairness. [2] Then in Section 3, we study representational biases in the underlying data sources and what they could imply for the extent of diversity that the tool can identify.

We analyse the two alternatives we are currently working with – DBpedia and Wikidata – with respect to their ontological coverage and diversity, and describe implications for the resulting analyses of text corpora. Specifically, we describe a case study that examines the relative over- or under-representation of Belgian political parties between 1990 and 2020 and highlights the many research-design decisions needed for this data analysis and the assumptions behind it, as well as implications from the results. In particular, we came across a staggering over-representation of the political right in the English-language DBpedia.

This result presents a fascinating challenge for the design of data- and AI-powered tools such as Diversity Searcher for public libraries. Through its base functionality, but maybe even more so through transparency regarding findings such as those in our case study, the tool can help foster fairness and diversity and create awareness among news users that multi-voiced news is necessary to form an accurate and fair view of the reality being reported in a news story.

# 2 The Diversity Searcher as a text analysis tool

The Diversity Searcher is a semi-automated text analysis and knowledge enrichment tool that is designed to present information to the user about the degree of diversity in news media texts.[3] The tool was developed in the interdisciplinary project DIAMOND (Diversity and Information Media: New Tools for a Multifaceted Public Debate)[4] to be used by media professionals and media consumers, and to be integrated with iCandid,[5] a research infrastructure project offering integrated access several (social)media resources to researchers. Over the course of the project, Diversity Searcher has been further developed to be a tool to be used in libraries as well.

The tool analyses an uploaded text or collection of texts. It extracts actors (persons, institutions, and geopolitical entities) and offers additional information and interaction opportunities about the individual actors and their occurrence(s) in the text(s). These range from base statistics (such as frequencies of different actors) to a numerical evaluation of the complex notion of "diversity".

---

[1] See for example Kitchin, Rob: The Data Revolution. Big Data, Open Data, Data Infrastructures & Their Consequences. London: Sage. 2014.

[2] Berendt, Bettina; Karadeniz, Özgür; Mertens, Stefan u. a.: Fairness beyond "equal": The Diversity Searcher as a Tool to Detect and Enhance the Representation of Socio-political Actors in News Media, in: Companion Proceedings of the Web Conference 2021, Ljubljana Slovenia 2021, S. 202–212. Online: <https://doi.org/10.1145/3442442.3452303>, Retrieved: 25.08.2022.

[3] Ibid.

[4] About DIAMOND, 07.01.2018, <https://soc.kuleuven.be/ims/diamond/diamond-about>, Retrieved: 27.08.2022.

[5] iCandid, A Snapshot of LIBIS Research Infrastructures, 03.05.2022, <https://bib.kuleuven.be/english/libis/projects#icandid>, Retrieved: 25.08.2022.

## 2.1 A quantitative measure of (actor) diversity in a text

One of our core assumptions is that diversity in media is a crucial precondition of a well-functioning democracy and a well-informed public opinion. Diversity is considered a crucial part of information quality, for which we draw on Stirling's (2007)[6] influential proposal of a quantitative measure that draws on a meta-analysis of the literature on diversity from various disciplines. Stirling identified three components, which all enhance diversity:

- **Variety**: many distinct entities are present, e.g. a number of people of different backgrounds;
- **Disparity**: the entities are different from one another, they come from a wide range;
- **Balance**: the different entities are evenly distributed. This component is a simple form of the "equality of ..." concept found in the fairness literature.

Based on these three components Stirling defined diversity Δ as

$$\Delta = \sum_{i,j(i \neq j)} (d_{ij})^\alpha \cdot (p_i \cdot p_j)^\beta$$

with variety being captured by the cardinality of the set of all entities i, j ∈E present in the domain (e.g. a text, a population, ...), balance by the frequencies $p_i$ (the more uniform the distribution, the higher this multiplicative factor), and disparity by a measure of distance or dissimilarity d(.,.). In addition, the parameters α and β allow for a weighting of the importance of balance or disparity.

Hence, diversity in news media means that many news stories are available (variety) and that these express different viewpoints (disparity), with all views equally represented (balance). As Ranaivoson points out, disparity is often a hurdle when using this formula, as it is measured based on strong assumptions.[7] The tool therefore focuses on a specific aspect to address diversity in media content: actor diversity.

## 2.2 An outline of the processing in the Diversity Searcher

In this subsection, we give a brief overview of the Diversity Searcher's processing. For details, user interface, and a worked-out example, see (Berendt et al., 2021).[8]

---

[6] Stirling, Andy: A general framework for analysing diversity in science, technology and society, in: Journal of The Royal Society Interface 4 (15), 22.08.2007, S. 707–719. Online: <https://doi.org/10.1098/rsif.2007.0213>.

[7] Ranaivoson, Heritiana: Measuring cultural diversity with the Stirling model, in: New Techniques and Technologies for Statistics, 2013, S. 10. Online: <https://ec.europa.eu/eurostat/cros/content/measuring-cultural-diversity-stirling-model-heritiana-ranaivoson_en>.

[8] Berendt, Bettina; Karadeniz, Özgür; Mertens, Stefan u. a.: Fairness beyond "equal": The Diversity Searcher as a Tool to Detect and Enhance the Representation of Socio-political Actors in News

We study the diversity of socio-political actors, understood broadly in terms of three types of actors having socio-political agency such as: persons, organisations and geo-political entities. Our design choices were inspired by a sociological understanding[9] of the extent and the ways in which two social actors are different.

We then built a knowledge retrieval and text processing pipeline that draws on data from an underlying knowledge source ontology. Originally, we worked with the English-language DBpedia. The Diversity Searcher performs its analyses in two steps. The first one consists of detecting the actors in the texts with a combination of Named Entity Recognition(NER), Named Entity Linking (NEL), and case sensitive and insensitive text and lemma matching.[10] For NEL, we use DBpedia Spotlight, a NEL model available in multiple languages, which links entities to DBpedia.[11] NER and pattern matching tasks use the third party tool spaCy, a freely available NLP library and language models.[12] The pattern matching function was added to the pipeline when our colleagues in media studies observed that Spotlight was performing poorly for Flemish politicians. The Addition of this feature also paved the way for user-defined recognition rules for named entities, and unnamed entities such as 'refugee', 'criminal', etc.

The application includes a corpus upload function. Using this function, users can upload XML files exported from Belga Press[13] and batch process them, as well as export the results as an Excel file to be used in further analysis in Excel or SPSS.

Although detection of the actors is sufficient to address balance and variety, disparity requires further information about the detected entities, hence the knowledge retrieval and enrichment stage. As DBpedia contains information that is both relevant and irrelevant to our task, the analysis also requires filtering the predicate-object pairs we obtain from DBpedia. For this task, we developed our own ontology consisting of socio-politically relevant actors and features. Graph data from DBpedia is thus transformed into an array of "feature_name": "feature_value" pairs in accordance with our ontology, which are inserted into the local record of the resource, and later used in calculations.

The knowledge retrieval and ontology mapping processes all the resources linked to the root resource, provided that they also belong to the three actor types. This decision is based on the fact that properties of some types of entities that are linked to the root entity also contain important information about the root entity. Thus, the Diversity Searcher does not stop after

---

Media, in: Companion Proceedings of the Web Conference 2021, Ljubljana Slovenia 2021, S. 202–212. Online: <https://doi.org/10.1145/3442442.3452303>, Retrieved: 25.08.2022.
[9] Bourdieu, Pierre: The Social Space and the Genesis of Groups. Theory and Society 14, 6. 1985. 723-744.
[10] NER: The subtask of Natural Language Processing (NLP), consisting in categorizing named entities in a text into predefined categories; NEL: The subtask of NLP consisting in linking named entities in a text to resources on a remote knowledge base.
[11] Mendes, Pablo N.; Jakob, Max; García-Silva, Andrés u. a.: DBpedia spotlight: shedding light on the web of documents, in: Proceedings of the 7th International Conference on Semantic Systems - I-Semantics '11, Graz, Austria 2011, S. 1–8. Online: <https://doi.org/10.1145/2063518.2063519>, Retrieved: 25.08.2022.
[12] Honnibal, Matthew; Montani, Ines: spaCy 2: Natural language understanding with Bloom embeddings, convolutional neural networks and incremental parsing, 2017. <https://spacy.io/>, Retrieved: 29.08.2022.
[13] Belga Press (belgapress.be) is an online press database covering a large number of international and Belgian news sources, where journalists and media researchers can search and export corpora. The specific source and therefore its XML format are relevant for our target users, but the schema could easily be made configurable.

retrieving, for example, the party of a politician: it will also attempt to retrieve the party's properties such as ideology, as different parties (variety) may have intersecting ideologies, thus affecting disparity. Similarly, finding the country of an actor will result in a further query into properties of the country such as EU membership or type of government.

Despite contributing to enrichment by further contextualising the actors, the integration of information from linked entities has the disadvantage of including these resources' errors and biases. For example, the data quality of the entry for "US" will also affect the data about "Hillary Clinton", "Joe Biden" and "Donald Trump" in the local knowledge base. Similarly, any systematic political bias related to representation of political parties of certain ideologies will affect the tool's representation of the root politician, even if they are detected correctly.

# 3 The role of the underlying ontology for bias and diversity: Comparing DBpedia and Wikidata

Tasks like that of the Diversity Searcher critically depend on external knowledge sources. In choosing which one(s) to use, one needs to compare different options. The question is how to do this in a formal and automated manner and in a way that supports the goals of the application. We opted for a method that is based on (a) a number of observations and assumptions about the available external sources, (b) the implementation of these assumptions into computational procedures that extract a relational representation from the RDF structures, and (c) application-specific questions about the data obtained. The criteria build on the comprehensive framework for determining data quality on Linked Open Data (LOD) by Zaveri et al. on the one hand and the literature on measuring bias in language (models) on the other hand, and they re-interpret these in the light of our questions around actor diversity.[14] [15]

For example, even if a news media article is highly diverse (say, in political affiliation of the represented actors), if the entity recognition process relies on a knowledge source that only contains actors of one political affiliation, only these actors will be processed in the diversity calculation and result in a minimal diversity value. So an ideal knowledge source would exhibit high coverage, correctness and timeliness, and high diversity.

We therefore began to question whether our initial choice of DBpedia was useful to debate what consequences the choice of the English-language DBpedia versus (if applicable) the language of the country or region under investigation would have (in general, local DBpedias have better coverage of local actors), and to suspect that Wikidata, which operates with more stringent quality controls,[16] could be a better basis for the Diversity Searcher.

---

[14] Amrapali Zaveri, Anisa Rula, Andrea Maurino, Ricardo Pietrobon, Jens Lehmann, Sören Auer: Quality assessment for Linked Data: A Survey. Semantic Web 7(1): 63-93 (2016)

[15] Delobelle, Pieter; Tokpo, Ewoenam; Calders, Toon u. a.: Measuring Fairness with Biased Rulers: A Comparative Study on Bias Metrics for Pre-trained Language Models, in: Proceedings of the 2022 Conference of the North American Chapter of the Association for Computational Linguistics: Human Language Technologies, Seattle, United States 2022, S. 1693–1706. Online: <https://doi.org/10.18653/v1/2022.naacl-main.122>, Retrieved: 25.08.2022.

[16] Piscopo, Alessandro; Simperl, Elena: What we talk about when we talk about wikidata quality: a literature survey, in: Proceedings of the 15th International Symposium on Open Collaboration, Skövde Sweden 2019, S. 1–11. Online: <https://doi.org/10.1145/3306446.3340822>, Retrieved: 29.08.2022.

These questions present two types of challenges for tool design. The first is pragmatic/implementation-related, and the second is conceptual. The selection of a knowledge source may appear, in the age of LOD, to be a modular and thus easily configurable design feature. Indeed choices, including flexible and/or user-determined choices,[17] between these knowledge sources can be implemented in a tool such as the Diversity Searcher. However, they require preparatory studies of these datasets and adaptations of the extraction algorithms, since predicate names and coding conventions differ (e.g., between DBpedia and Wikidata) and because even within DBpedia, the possible universality of predicates is not always used and different language versions exhibit preferential uses of different predicates with related but not equal syntax and semantics. For example, the possibility of encoding the temporal dimensions of politicians' active terms is used widely in the English DBpedia but not in the other sources, where instead positions held are often specified. It is possible that such differences result from authors in one language DBpedia-authors community copying the information-presentation style of others in their own community. In addition, in each case there are reasons for tool builders to prefer uniformity (the extraction of the same predicates) or to choose predicates based on other criteria. We opted for the latter approach and chose, from semantically highly related predicates, those with the highest coverage.

## 3.1 Basic statistics

We first considered the English DBpedia and Wikidata as the two external knowledge sources; the method can easily be generalised to others. Being the two largest sources of LOD, both DBpedia and Wikidata have the potential to provide relevant information for the entity types that Diversity Searcher recognizes.[18] [19] [20] Although both contain machine-readable, freely available data based on Wikipedia, they differ in their ontologies and construction.

DBpedia is the older of the two and is defined as "a crowdsourced community effort to extract structured content from the information created in various Wikimedia projects."[21] DBpedia extracts information from Wikipedia pages (mainly from the info-boxes) as RDFs and creates unique (language-dependent) URIs for its resources from the pages in a semi-automatic way.[22] This way, it forms an open language graph that is integrated well within the Semantic Web and popular among the LOD communities.[23]

---

[17] Combinations of several knowledge sources are also possible, but require even more choices, especially in cases when there is conflicting information in two sources. This question will therefore be left for future work.

[18] Auer, Sören; Bizer, Christian; Kobilarov, Georgi u. a.: DBpedia: A Nucleus for a Web of Open Data, in: Aberer, Karl; Choi, Key-Sun; Noy, Natasha u. a. (Eds.): The Semantic Web, Bd. 4825, Berlin, Heidelberg 2007 (Lecture Notes in Computer Science), S. 722–735. Online: <https://doi.org/10.1007/978-3-540-76298-0_52>

[19] Lehmann, Jens; Isele, Robert; Jakob, Max u. a.: DBpedia - A Large-scale, Multilingual Knowledge Base Extracted from Wikipedia, in: Semantic Web Journal 6 (2), 2015, S. 167–195.

[20] Vrandečić, Denny; Krötzsch, Markus: Wikidata: a free collaborative knowledgebase. In: Communications of the ACM 57 (10), 2014, pp. 78–85. Online: <http://dx.doi.org/10.1145/2629489>.

[21] About DBpedia, DBpedia, (n.d.).

Wikidata differs from DBpedia in many ways. First, they relate to Wikipedia differently. Wikidata aims to provide structured data for all other Wikimedia projects and contains entities that are not covered in Wikipedia. However, DBpedia is strictly limited by the information available on Wikipedia. Second, Wikidata is edited and maintained by a large and open community of contributors, who fact-check and enrich the data via various WikiProjects.[24] DBpedia works based on a closed community and depends more on the correctness of information on Wikipedia. Third, the Wikidata dataset is not language-dependent and consists of one large graph that can be displayed using any available language label or description. Unlike the language-dependent DBpedia datasets, URIs are generated as P-numbers for properties and Q-numbers for entities in Wikidata. Fourth, thanks to its RDF structure Wikidata allows qualifiers and references to be added to the statements, and this extra information can be helpful in various NLP tasks. Finally, while both offer SPARQL APIs, Wikidatahas more restrictions and is less suitable for extracting large amounts of data.[25]

DBpedia and Wikidata also differ in the amount of information they offer. The latest snapshot of the English version of DBpedia contains more than 850 million triples describing about 5.4 million entities.[26] Wikidata, on the other hand, provides about 1.36 billion triples describing about 97 million entities.[27] [28] Consequently, the average number of statements per entity is approximately 157 in DBpedia and about 14 in Wikidata. Thus, as previous research indicated, Wikidata describes more entities, while DBpedia has more statements about each entity.[29] While DBpedia offers various statements in many semantic languages about its entities, Wikidata offers more resources with more structured and compact information that offers higher utility for certain specific use cases. This summarization of findings in the literature is also reflected by our own results, which are to be presented next.[30]

---

[22] Hofer, Marvin; Hellmann, Sebastian; Dojchinovski, Milan u. a.: The New DBpedia Release Cycle: Increasing Agility and Efficiency in Knowledge Extraction Workflows, in: Blomqvist, Eva; Groth, Paul; Boer, Victor de u. a. (Eds.): Semantic Systems. In the Era of Knowledge Graphs, Bd. 12378, Cham 2020 (Lecture Notes in Computer Science), S. 1–18. Online: <https://doi.org/10.1007/978-3-030-59833-4_1>, Retrieved: 21.02.2022.

[23] Abián, D.; Guerra, F.; Martínez-Romanos, J. u. a.: Wikidata and DBpedia: A Comparative Study, in: Szymański, Julian; Velegrakis, Yannis (Eds.): Semantic Keyword-Based Search on Structured Data Sources, Bd. 10546, Cham 2018 (Lecture Notes in Computer Science), S. 142–154. Online: <https://doi.org/10.1007/978-3-319-74497-1_14>, Retrieved: 25.08.2022.

[24] Wikidata:Statistics, <https://www.wikidata.org/wiki/Wikidata:Statistics>, Wikidata, (n.d.), Retrieved: 01.03.2022.

[25] The current limits of the APIs can be found in the following links:
<https://www.dbpedia.org/resources/sparql/>;
<https://www.mediawiki.org/wiki/Wikidata_Query_Service/User_Manual>

[26] Holze, Julia: DBpedia Snapshot 2021-12 Release Announcement, DBpedia Blog, 09.02.2022, <https://www.dbpedia.org/blog/snapshot-2021-12-release/>, Retrieved: 28.08.2022.

[27] Wikidata dashboards / Wikidata Datamodel Statements, Grafana, (n.d.), <https://grafana.wikimedia.org/d/000000175/wikidata-datamodel-statements?orgId=1&refresh=30m>, Retrieved: 01.03.2022.

[28] Wikidata:Statistics, (n.d.)

[29] Abián, et al.: Wikidata and DBpedia: A Comparative Study, 2018

## 3.2 A brief exploration of coverage of the domain of politicians

In a first data exploration, we focused on examples of the coverage of politicians in the English DBpedia and in Wikidata, due to their relevance for the Diversity Searcher. We formulated the following SPARQL queries:

| Query | English-language DBpedia | Wikidata |
| --- | --- | --- |
| Members of the Belgian Chamber of Representatives [31] | 143 | 2996 |
| Members of the Flemish Parliament [32] | 21 | 464 |
| Politicians who held office in the US Chamber of Representatives [33] | 14886 | 11,160 |

Table 1. Three queries to compare the coverage of politicians in the English-language DBpedia and Wikidata.

Looking into the results of the second query, we noticed that 15 (about ¾) of the provided Flemish politicians belong to one political party (N-VA, Nieuw-Vlaamse Alliantie, New Flemish Alliance). Although N-VA is currently the largest party in the Flemish parliament with 35 seats out of 124, their members are overrepresented in English DBpedia compared to other Flemish parties.[34] The data also showed some miscategorizations in DBpedia.[35]

---

[30] Färber, Michael, Frederic Bartscherer, Carsten Menne, and Achim Rettinger. 'Linked Data Quality of DBpedia, Freebase, OpenCyc, Wikidata, and YAGO'. Edited by Amrapali Zaveri, Dimitris Kontokostas, Sebastian Hellmann, Jürgen Umbrich, Amrapali Zaveri, Dimitris Kontokostas, Sebastian Hellmann, and Jürgen Umbrich. Semantic Web 9, no. 1 (30 November 2017): 77–129. https://doi.org/10.3233/SW-170275. pp.46-8

[31] The link to the DBpedia page: https://dbpedia.org/page/Category:Members_of_the_Chamber_of_Representatives_(Belgium) and the Wikidata query: https://w.wiki/4DmS . However, we should also point out that the same list of members of the Belgian Chamber of Representatives in Dutch-Flemish DBpedia consists of 701 resources: http://nl.dbpedia.org/page/Kamer_van_volksvertegenwoordigers .Date of Access: 01/10/2021. We will look deeper into representational problems in each of these three databases further in the following sections.

[32] The link to the DBpedia query: https://tinyurl.com/4uwbycwu and the Wikidata query: https://w.wiki/3YG4. Date of Access: 01/10/2021.

[33] The link to the DBpedia query: https://tinyurl.com/e3y2b8kh Date of access and the Wikidata query: https://w.wiki/4DnP. Date of Access: 01/10/2021.

These results indicate that the English DBpedia may offer fewer entities (especially in non-English language contexts) and that there appear to be issues in terms of bias and quality. Our observations comply with the existing comparative literature concerning these two information sources. [36]

### 3.3 A deeper look at the representation of political parties

#### 3.3.1 Motivation: Have right-wingers taken over Wikipedia/DBpedia? And why this shows we need to look at diversity *and* bias

To understand the possible representational bias in DBpedia better, we followed up with a closer inspection of such biases and what they could mean for the Diversity Searcher's notion of diversity. In particular, we asked how we could capture the biases created by "over-representation", and how "over-presentation in a source ontology" should be defined in the context of diversity search in the first place. Given that political analyses always need to be understood in their historical context, in addition we asked what over-representation of (also) historical facts with a knowledge source that has a current and ever-evolving status could mean.

These questions also become an investigation into DBpedia, Wikidata, etc. as an infrastructure of cultural memory. In the case investigated here, that of representation of politicians by party, it could be argued that normative, correct and comprehensive sources probably exist in the form of parliament records, political party records, etc. and that the use of DBpedia/Wikidata is primarily a pragmatic and money-saving choice; but of course the question of representation and the needed input for the Diversity Searcher range beyond politicians and party affiliation, such that crowdsourced sources with all their imperfections are often the only forms of cultural memory. Thus, the investigation stands between the consequences of a pragmatic choice of source and the study of that source in itself.

#### 3.3.2 Method

The ensuing analysis proceeded through six stages.

1. We concentrated on the **attributes** "political party affiliation" and "political alignment of a party" as the feature on which assessments of bias and diversity are based. We limited our analysis to "politicians from Belgium", since we realised early on that substantial domain knowledge is needed to understand even purely numerical phenomena and members of our team are experts in this domain.

---

[34] The party membership information is gathered manually based on the previously mentioned DBpedia query (https://tinyurl.com/4uwbycwu). Date of Access: 01/10/2021.
[35] For example, DBpedia considers the Women's Equality Party (New York) as a type of "person" and lists it as a politician: https://dbpedia.org/page/Women%27s_Equality_Party_(New_York)
[36] Abián, et al: Wikidata and DBpedia: A Comparative Study, 2018

2. As a **null hypothesis**, a knowledge source represents a political constellation in an unbiased way if the relative number of politicians from a given party who are represented as an entity in a knowledge source[37] equals the relative number of this party in a relevant real-life context. We chose relative numbers because it was clear from the outset that our knowledge sources are not normative and comprehensive public records, and because we consider "having a Wikipedia page" (etc.) as an important contributor to public visibility of a person and their party.

This general question is only meaningful at a given point in time, since the baseline evolves, at the very least with elections that change political representation.

3. The **baseline** is then – relatively – easy to define: the shares of the vote or the number of seats of parties Y at times T in a given political body. We started by concentrating on the national parliament, the Chamber of People's Representatives (*Kamer van volksvertegenwoordigers*, henceforth *KVV*) and used the number of seats at the beginning of a legislature. We also looked at the regional (Flemish) parliaments (*Vlaams parlement, VP*), whose election time points are not identical to those at the national level.

These data were retrieved from *Wikipedia, de vrije encyclopedie* (2022a, 2022b).[38] [39]

4. The **representation in the ontology** is more difficult to define. One may want to compare parties' share at T (step 3.) with what was represented in the ontology at T. We chose to compare with what is represented in the ontology "now, but as relevant at T".  This has two reasons. First, we wanted to study the historical evolution since 1990, and Wikipedia only started operations at the beginning of 2001 and after that took some time to evolve, in particular in its non-English versions (for example, the Dutch-language version was created in 2001). Wikidata was launched in October 2012. Second, we aimed at studying these ontologies also as infrastructures of cultural memory, which changes the question slightly: How does a user currently see the political landscape (of the 1990s, etc.) through the lens of an ontology?

CSV files of actors and parties "represented now" were created using SPARQL queries.from the ontology current at the time of writing.[40] The filtering as "relevant at T" happened in the following step 5.

---

[37] In the case of DBpedia, which is derived from Wikipedia, this corresponds to "a person having a Wikipedia page".

[38] Wikipedia-bijdragers: Kamer van volksvertegenwoordigers — Wikipedia, de vrije encyclopedie, 2022. Online: <https://nl.wikipedia.org/w/index.php?title=Kamer_van_volksvertegenwoordigers&oldid=61849646>, Retrieved: 25.08.2022.

[39] Wikipedia-bijdragers: Vlaams Parlement — Wikipedia, de vrije encyclopedie, 2022. Online: <https://nl.wikipedia.org/w/index.php?title=Vlaams_Parlement&oldid=61662474>, Retrieved: 25.08.2022.

[40] In line with the Diversity Searcher's queries to online ontologies, we initially designed our queries to acquire information for Belgian politicians as well as contextual information about their parties and countries. Details on the queried properties are described and motivated in (Berendt et al., 2021) and can be seen in the hyperlinked queries. Some of this information was not used in the case study's analysis because it was irrelevant to the question (e.g. a politician's gender, or country information). Some variations between queries and workarounds were necessary to accommodate differences

5. **Computing party representation shares.** How do we then get from "politician X exists in the ontology" to "party Y covers Z% of the representation of time T"?

5a. We first limited the analysis to politicians who were active at T. We derived this by creating each actor's maximal activity period as the union of all activity periods represented in the ontology, limited – in the absence of an explicit end date – by "today" or, if applicable, the actor's death date or, in a small number of cases, a well-publicised[41] date of ending their political career. We selected the politicians who were active at T = 1 January of 1990, 1996, 2000, 2005, 2011, 2015, 2020. (The exceptions from the 5-year spacing were done to capture the effects of the general elections held in 1995 and 2010.)

5b. We then pre-processed each actor's list of party affiliations (which was often long and heterogeneous). First, parties that changed their name at one or more time points were mapped to one, identified by their current acronym. Second, small parties were grouped as "not relevant", "foreign party", etc. and not further analysed due to their low shares. These mappings were created by the domain expert in our team.

5c. Third, this still left many politicians who belonged to different parties at different times in their career. At first sight, this may appear to be an instance of the "valid time" problem of temporal databases that can be handled by all state-of-the-art databases. Valid time is the time period during which a database fact is valid in the modelled reality. In principle, a tool that has identified an actor in a media text[42] could look up what party P the actor belonged to at the relevant time T and regard them as a representative of P in that context. However, what is that period when it comes to assessing a text? First, which time period(s) should count: when the article was written, what it describes, when it was read? Second, how should the period that the text talks about be recognised and delimited? Third, is the (factual or perceived) "identity" of an actor not shaped by their whole history? Therefore, rather than attempting to identify and process a valid time from the database facts, the Diversity Searcher implicitly assumes the perspective of "looking from the now" and thereby also imitate a human user.

But what would that imitated human user "see"? Looking from the now, the politician may be perceived as giving visibility to only one of their parties or to all of them. If they give visibility to all of them, they may do so equally, or one of their affiliations may "eclipse" the others. It could even be the case that these historical changes have focussed the attention only on

---

between the ontologies, their data quality, and technical restrictions on maximal number of results. The limitations of these pragmatic choices are discussed in Section 6.
The queries can be found at https://tinyurl.com/ys2bnyy7, https://tinyurl.com/ys2bnyy7, https://tinyurl.com/4ejp5jne, https://tinyurl.com/33fej4aw, https://tinyurl.com/2a3ph5bt (English DBpedia); https://w.wiki/56Qc, https://w.wiki/53Lb, https://w.wiki/53ap, https://w.wiki/53gw (Wikidata); https://tinyurl.com/2bjxp2f7, https://tinyurl.com/y5fzpu6p, https://tinyurl.com/56xkpy6m, https://tinyurl.com/38krrb57https://tinyurl.com/38krrb57 (Dutch DBpedia). Dates of Access (respectively): 27.05.2022, 27.05.2022, 12.04.2022, 27.05.2022, 12.04.2022, 13.04.2022, 12.04.2022, 13.04.2022, 26.04.2022, 13.04.2022, 12.04.2022, 12.04.2022. Python script used as a workaround for the query limitations are provided in an additional document.
[41] We used Wikipedia as knowledge source to determine these cases.
[42] See Section 2 for a brief description of how Diversity Searcher identifies and processes actors.

that person and obfuscated that they also belonged to a party. Which of these apply for a given politician and a given time point, is likely to depend heavily on the politician, the time point, the political actions and narratives created by and around the parties[43], the user, and probably other factors. In the absence of the psychological empirical work needed to determine which of these cases apply, we opted for delimiting the options by bounds on the representation:

A party is made visible at time T *at least* by all those politicians who, over their whole career, were *only* in this party[44]. This number defines a lower bound its visibility at T, because it assumes that anyone who was in multiple parties does not confer any visibility on any of them. The party is made visible *at most* by all those politicians who were ever in it at any time in their career. This number defines an upper bound on the party's visibility at T, because it assumes that anyone who was in multiple parties confers full visibility on all of them. All settings in which different multi-party politicians confer more or less visibility to their various parties, lie between these bounds.

These bounds are then compared to the baseline shares. If the lower bound is above the baseline, the party is certainly over-represented. (Ex.: a party has 20% of the vote, but 30% of all politicians in the ontology are pure-bred members of this party, and there may be additional others who were in it at some point.) If the upper bound is below the baseline, it is certainly under-represented. (Ex.: a party has 20% of the vote, but 10% of all politicians in the ontology were in that party at all … and maybe in others too.) When the baseline lies between lower and upper bound, no clear conclusion can be drawn about over- or under-representation.

Step 5 involved post-processing and querying in a MySQL relational database into which the CSV files generated in Step 4 were imported.

6. **Studying possible over-/under-representations and their trends over time.** The analysis proceeded descriptively and via the visualisations shown in the following section. In principle, we should clarify "higher" and "lower" in statistical terms (both in terms of statistical significance and effect size). However, given the small number of actors and an as-yet-absent knowledge of distributions, we limit ourselves to descriptive statistics.

### 3.3.3 Results and interpretation

The results are summarised in Figure 1. The figure highlights the *trends* in the data and the *contrast* between the knowledge sources, in terms of an aligned sorting of parties from left to right and curves lying consistently above or below the baseline over time. *Details* can be seen in the higher-resolution/interactive online versions.[45] Bold lines above thin lines indicate an over-representation, bold lines below thin lines indicate an under-representation of that party and in this sense also of the political alignment it is part of. The English-language DBpedia has less than 20 actors up until 2005, which makes the over- and under-representation hard to interpret. The number of politicians represented in the Dutch-language DBpedia fell sharply after the last regional and national elections in 2019, making also this graph harder to interpret relative to the others.

---

[43] such as rapprochements between right-wing and extreme-right parties that lead to the boundaries between them becoming blurred
[44] We disregard memberships in a "not relevant" minor party with negligible votes.
[45] http://www.berendt.de/DIAMOND/, Retrieved: 27.12.2022.

These results not only confirm our first informal observation of over-representation of right-wing parties (especially the N-VA) in the English-language DBpedia, with a trend growing over time. (During these years, the N-VA's share of the popular vote increased, but the DBpedia growth clearly exceeds the baseline growth.) Different biases seem to occur in the Dutch-language DBpedia: although on the whole comparatively similar to the baseline, this ontology seems to over-represent the main centrist party (CD&V). Wikidata, in contrast, gives a rather accurate picture of party shares in the national parliament. The French-language Walloon parties are (understandably, given the language focus) under-represented in the Dutch-language DBpedia. Both the overrepresentation of rightist and centrist parties in media coverage have been identified in earlier international research, such as a centrist bias in the media coverage of the UK elections of 2017[46] and the right-wing overrepresentation in social media, despite the cries of censorship in the United States.[47]

INSERT FIGURE 1 ABOUT HERE

The seeming over-representation of the political centre in the Dutch-language DBpedia may be an artefact of the language: The Dutch DBpedia focuses more on Flemish actors than on the French-speaking Walloon (or Brussels or German Community) actors, and an informal inspection of politicians' pages in this ontology indicated that many regional and local political actors are represented. Therefore, Figure 2 maps the shares of seats in the Flemish parliament (which was first elected directly in 1995) as the baseline and is otherwise analogous to Figure 1. The set of parties is a subset of that of Figure 1, since only the Flemish parties can be elected into the Flemish parliament (while the national parliament also contains representatives from the Walloon, Brussels, and German-Community parties). Figure 2 suggests that the representation of the parties between extreme left and centre-right in the Dutch-language DBpedia actually mirrors these parties' shares in the regional parliament rather closely; while the right and extreme right tend to be under-represented especially in the later years, when they were highly successful especially in the Flemish elections.

INSERT FIGURE 2 ABOUT HERE

---

[46] Deacon, David; Downey, John; Smith, David, Stanyer, James and Wring, Dominic. National News Media Coverage of the 2017 election. Centre for Research in Communication and Culture, Loughborough University Report 4: 5 May – 7 June 2017 Centre for Research in Communication and Culture, Loughborough University Online: <https://blog.lboro.ac.uk/crcc/wp-content/uploads/sites/23/2017/06/media-coverage-of-the-2017-general-election-campaign-report-4.pdf>

[47] Scott, Mark: Despite cries of censorship, conservatives dominate social media, POLITICO, 26.10.2020, <https://www.politico.com/news/2020/10/26/censorship-conservatives-social-media-432643>, Retrieved: 28.08.2022.

Given the low shares of small-party representatives both in the baselines and in the ontologies and given that our focus is on Flemish parties, we do not investigate the small or the Walloon parties' representations further.

### 3.3.4 Implications for the Diversity Searcher and future work

The political-science literature has hinted at the political right being skilled at using social media before, and this appears to be an international phenomenon. As regards the "over-representation of the centre", this could be an artefact of the Dutch-language resource's focus on Dutch-language (Flemish) politicians and on majorities on the regional level, an interpretation that is supported by the disappearance of the over-representation when comparing with the baseline of the Flemish parliament.

Why do such biases matter in the context of the Diversity Searcher? The tool can only compute diversity between entities that it recognises. Imagine (a) a well-balanced text with (say) several left-wing actors and several actors from one right-wing party, and (b) a text that only contains actors from the right-wing and extreme-right parties. Assuming that all else is equal between these texts, the "ground truth" would be that (a) is more diverse than (b). However, when sourced by an ontology with a bias to the right, the tool will only (or mostly) recognise actors from the one right-wing party in (a) and most actors from the two parties in (b), which will lead to a higher computed diversity score for text (b) and a lower score – maybe even zero – for text (a). This result is related to the artefact of ontological knowledge (also observed in humans) that one perceives more distinctions and diversity in areas that one knows well.

The case study also illustrated the non-trivial and multiple dependencies between diversity, fairness and biases. The feature "party" is one of the features from which the Stirling *disparity* between recognised entities is computed. As a result, a text's calculated *diversity* is influenced by this feature. In general, diversity increases with the *balance* of the different values of this feature (i.e., different parties). However, if these values' shares diverge too much from the baseline, the knowledge source is *biased* – and this in turn may bias the calculation of the diversity of the text, introducing *unfairness*.

Wikidata, by contrast, mirrors the representation of parties in the Belgian Parliament quite accurately. In future work, data quality should be investigated also for other relevant domains (such as other countries' party systems or other socio-politically relevant topics), in order to provide an evidence-based reason to choose one ontology over another for use in a tool such as the Diversity Searcher.

# 4 Limitations, conclusions and outlook

As Leorke et al. argue, the shifts towards digital economies and smart cities resulted in a change in the societal role of public libraries.[48] The latter are now increasingly considered and expected to function as "hubs" or "platforms" that "link people to information, services and to each other".[49] Moreover, public libraries also have the task of bridging digital divides by facilitating access to digital resources and training the public in digital literacy. This renewed societal role of the public libraries within the digital economy makes their increasing usage of automated content analysis more problematic. Automated content analysis software has the potential to provide insight into large digital corpora quickly, making them valuable additions to research and critical thinking in a library setting. However, as illustrated by the case study in this article, such tools are prone to reproducing bias in the upstream components, even when they are designed to alleviate biases. Working with data requires knowledge and recognition that data are not neutral and that they can be used to maintain an unequal status quo. Thus, data are part of the problem but can also be part of the solution if we maintain a number of principles: one is to situate the data in a historical and social context. Library scientists can help create this awareness and a critical and informed attitude at the user end.

---

[48] Leorke, Dale; Wyatt, Danielle; McQuire, Scott: "More than just a library": Public libraries in the 'smart city', in: City, Culture and Society 15, 12.2018, S. 37–44. Online: <https://doi.org/10.1016/j.ccs.2018.05.002>.

One limitation of the study was the inability to use the exact same SPARQL queries for each database due to Wikidata, English and Dutch versions of DBpedia all having differences in their ontologies and data quality. For some cases we found that using the same query texts or analogous properties favours one database while bringing an empty or limited result for another despite data being available.[50] As our research and tool development focuses on media research and library contexts, we prioritised acquiring the relevant data, even if it meantmeans using different SPARQL queries tailored for each database.

In future work, we plan to develop our analysis of biases in the underlying ontologies further. This includes (a) combinations of several knowledge sources (rather than exclusive choices between them), which requires further design decisions, especially in cases when there is conflicting information in two sources; and (b) an investigation of Wikipedia's edit history to complement the "view from now" by "recreating the knowledge at time point X in the past". We also aim at further developing the understanding (including formalisation) of the relationships between diversity, bias and fairness.

In general, our findings indicate that as developers of tools, we need to monitor the potential biases in our knowledge sources and study how they may influence the "downstream-task" results (such as the calculation and presentation of diversity) and also user perception. Apart from this, the authoring processes that lead to such over-representations are a highly interesting topic in its own right. In future work, we will study both questions and also relate them to our work on bias in large language models.[51]

It should not be forgotten that these numerical considerations can be exacerbated by the inability of automated tools to recognise and understand context. This is another reason to treat the numerical and categorical results as a *starting point* for deeper text analysis and involve users – whether it is an individual citizen, researcher, organisation representing a particular target group or a journalist looking for a new angle for a news story – in sense-making by means of the interactive interface of the Diversity Searcher. This idea can and should be extended. For future work, a two-pronged strategy is recommended: (a) identifying and using the best-suited ontology for a given task and at the same time (b) making its properties and shortcomings transparent to users so as to keep users aware of challenges associated with the (and any) dataset.

---

[49] Ibid.
[50] For example, the query for entities of type 'political party' and with country 'Belgium', which we used for English DBpedia and Wikidata, returned no results for Dutch DBpedia even though the entities exist. As a workaround, we queried the Dutch DBpedia for entities that appear as the party property of another entity, and that have country 'Belgium'.
[51] Delobelle, Tokpo, Calders, u. a.: Measuring Fairness with Biased Rulers, 2022

# Acknowledgements

We thank the Fonds Wetenschappelijk Onderzoek – Vlaanderen (FWO) for funding DIAMOND under project code S008817N.

# References

About DBpedia, DBpedia, (n.d.)., <https://www.dbpedia.org/about/>, Retrieved: 28.08.2022.

About DIAMOND, 07.01.2018, <https://soc.kuleuven.be/ims/diamond/diamond-about>, Retrieved: 27.08.2022.

Abián, D.; Guerra, F.; Martínez-Romanos, J. u. a.: Wikidata and DBpedia: A Comparative Study, in: Szymański, Julian; Velegrakis, Yannis (Eds.): Semantic Keyword-Based Search on Structured Data Sources, Bd. 10546, Cham 2018 (Lecture Notes in Computer Science), S. 142–154. Online: <https://doi.org/10.1007/978-3-319-74497-1_14>, Retrieved: 25.08.2022.

Amrapali Zaveri, Anisa Rula, Andrea Maurino, Ricardo Pietrobon, Jens Lehmann, Sören Auer: Quality assessment for Linked Data: A Survey. Semantic Web 7(1): 63-93 (2016)

Auer, Sören; Bizer, Christian; Kobilarov, Georgi u. a.: DBpedia: A Nucleus for a Web of Open Data, in: Aberer, Karl; Choi, Key-Sun; Noy, Natasha u. a. (Eds.): The Semantic Web, Bd. 4825, Berlin, Heidelberg 2007 (Lecture Notes in Computer Science), S. 722–735. Online: <https://doi.org/10.1007/978-3-540-76298-0_52>

Berendt, Bettina; Karadeniz, Özgür; Mertens, Stefan u. a.: Fairness beyond "equal": The Diversity Searcher as a Tool to Detect and Enhance the Representation of Socio-political Actors in News Media, in: Companion Proceedings of the Web Conference 2021, Ljubljana Slovenia 2021, S. 202–212. Online: <https://doi.org/10.1145/3442442.3452303>, Retrieved: 25.08.2022.

Bourdieu, Pierre: The Social Space and the Genesis of Groups. Theory and Society 14, 6. 1985. 723-744.

Deacon, David; Downey, John; Smith, David, Stanyer, James and Wring, Dominic. National News Media *Coverage of the 2017 election. Centre for Research in Communication and Culture, Loughborough University Report 4: 5 May – 7 June 2017* Centre for Research in Communication and Culture, Loughborough University Online: <https://blog.lboro.ac.uk/crcc/wp-content/uploads/sites/23/2017/06/media-coverage-of-the-2017-general-election-campaign-report-4.pdf>

Delobelle, Pieter; Tokpo, Ewoenam; Calders, Toon u. a.: Measuring Fairness with Biased Rulers: A Comparative Study on Bias Metrics for Pre-trained Language Models, in: Proceedings of the 2022 Conference of the North American Chapter of the Association for Computational Linguistics: Human Language Technologies, Seattle, United States 2022, S. 1693–1706. Online: <https://doi.org/10.18653/v1/2022.naacl-main.122>, Retrieved: 25.08.2022.

Färber, Michael, Frederic Bartscherer, Carsten Menne, and Achim Rettinger. 'Linked Data Quality of DBpedia, Freebase, OpenCyc, Wikidata, and YAGO'. Edited by Amrapali Zaveri, Dimitris Kontokostas, Sebastian Hellmann, Jürgen Umbrich, Amrapali Zaveri, Dimitris Kontokostas, Sebastian Hellmann, and Jürgen Umbrich. Semantic Web 9, no. 1 (30 November 2017): 77–129. https://doi.org/10.3233/SW-170275. pp.46-8

Hofer, Marvin; Hellmann, Sebastian; Dojchinovski, Milan u. a.: The New DBpedia Release Cycle: Increasing Agility and Efficiency in Knowledge Extraction Workflows, in: Blomqvist, Eva; Groth, Paul; Boer, Victor de u. a. (Hg.): Semantic Systems. In the Era of Knowledge Graphs, Bd. 12378, Cham 2020 (Lecture Notes in Computer Science), S. 1–18. Online: <https://doi.org/10.1007/978-3-030-59833-4_1>, Retrieved: 21.02.2022.

Holze, Julia: DBpedia Snapshot 2021-12 Release Anouncement, DBpedia Blog, 09.02.2022, <https://www.dbpedia.org/blog/snapshot-2021-12-release/>, Retrieved: 28.08.2022.

Honnibal, Matthew; Montani, Ines: spaCy 2: Natural language understanding with Bloom embeddings, convolutional neural networks and incremental parsing, 2017. <https://spacy.io/>, Retrieved: Retrieved: 29.08.2022.

iCandid, A Snapshot of LIBIS Research Infrastructures, 03.05.2022, <https://bib.kuleuven.be/english/libis/projects#icandid>, Retrieved: 25.08.2022.

Kitchin, Rob: The Data Revolution. Big Data, Open Data, Data Infrastructures & Their Consequences. London: Sage. 2014.

Lehmann, Jens; Isele, Robert; Jakob, Max u. a.: DBpedia - A Large-scale, Multilingual Knowledge Base Extracted from Wikipedia, in: Semantic Web Journal 6 (2), 2015, S. 167–195.

Leorke, Dale; Wyatt, Danielle; McQuire, Scott: "More than just a library": Public libraries in the 'smart city', in: City, Culture and Society 15, 12.2018, S. 37–44. Online: <https://doi.org/10.1016/j.ccs.2018.05.002>.

Mendes, Pablo N.; Jakob, Max; García-Silva, Andrés u. a.: DBpedia spotlight: shedding light on the web of documents, in: Proceedings of the 7th International Conference on Semantic Systems - I-Semantics '11, Graz, Austria 2011, S. 1–8. Online: <https://doi.org/10.1145/2063518.2063519>, Retrieved: 25.08.2022.

Piscopo, Alessandro; Simperl, Elena: What we talk about when we talk about wikidata quality: a literature survey, in: Proceedings of the 15th International Symposium on Open

Collaboration, Skövde Sweden 2019, S. 1–11. Online: <https://doi.org/10.1145/3306446.3340822>, Retrieved: 29.08.2022.

Ranaivoson, Heritiana: Measuring cultural diversity with the Stirling model, in: New Techniques and Technologies for Statistics, 2013, S. 10. Online: <https://ec.europa.eu/eurostat/cros/content/measuring-cultural-diversity-stirling-model-heritiana-ranaivoson_en>.

Scott, Mark: Despite cries of censorship, conservatives dominate social media, POLITICO, 26.10.2020, <https://www.politico.com/news/2020/10/26/censorship-conservatives-social-media-432643>, Retrieved: 28.08.2022.

Stirling, Andy: A general framework for analysing diversity in science, technology and society, in: Journal of The Royal Society Interface 4 (15), 22.08.2007, S. 707–719. Online: <https://doi.org/10.1098/rsif.2007.0213>.

Wikipedia-bijdragers: Kamer van volksvertegenwoordigers — Wikipedia, de vrije encyclopedie, 2022. Online: <https://nl.wikipedia.org/w/index.php?title=Kamer_van_volksvertegenwoordigers&oldid=61849646>, Retrieved: 25.08.2022.

Wikidata dashboards / Wikidata Datamodel Statements, Grafana, (n.d.), <https://grafana.wikimedia.org/d/000000175/wikidata-datamodel-statements?orgId=1&refresh=30m>, Retrieved: 01.03.2022.

Wikidata: Statistics, Wikidata, (n.d.), <https://www.wikidata.org/wiki/Wikidata:Statistics>, Retrieved: 01.03.2022

Vrandečić, Denny; Krötzsch, Markus: Wikidata: a free collaborative knowledgebase. In: Communications of the ACM 57 (10), 2014, pp. 78–85. Online: <http://dx.doi.org/10.1145/2629489>.

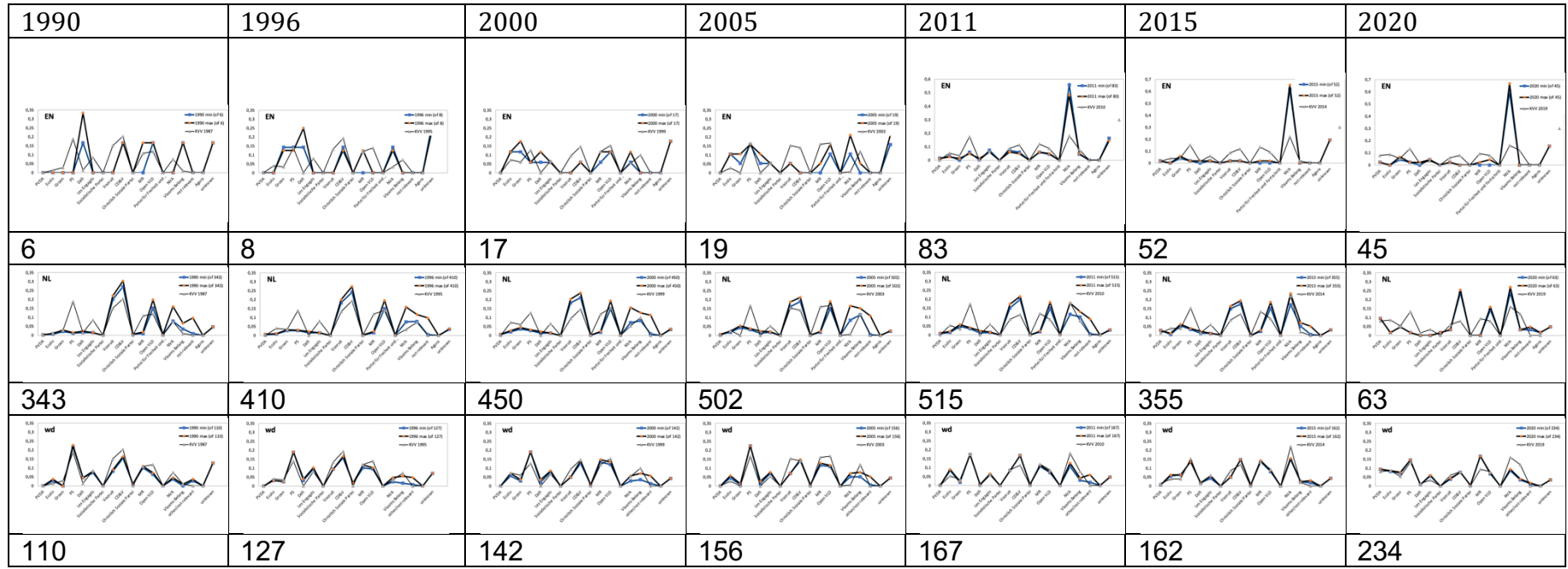

Figure 1. Visibility of parties in the English-language DBpedia (top), the Dutch-language DBpedia (middle) and Wikidata (bottom) from 1990-2020. Parties are ordered by alignment category (extreme left, left, center-left, center, center-right, right, extreme right, other, unknown) and within these categories by their current acronym. Bold lines or vertical spaces between bold lines are the shares in the knowledge source, the thin lines are the closest-in-time national parliament shares of the vote. The numbers below the graphs are the numbers of unique actors "active at the beginning of the year" in that knowledge source..

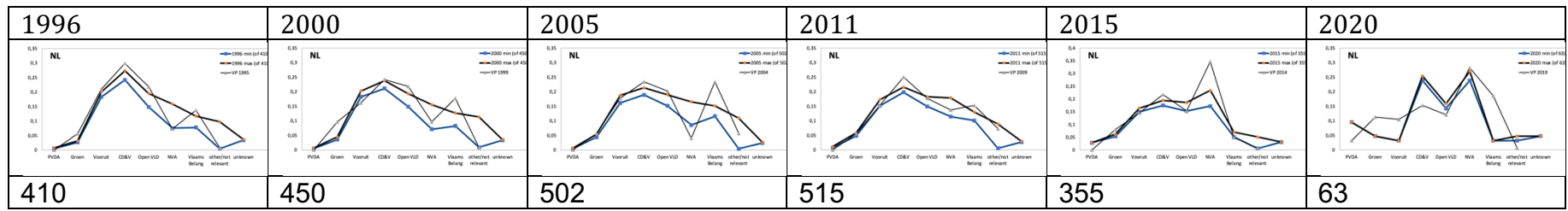

Figure 2. Visibility of Flemish parties in the Dutch-language DBpedia from 1996-2020. Parties are ordered by alignment category (extreme left, left, centre-left, centre, centre-right, right, extreme right, other, unknown) and within these categories by their current acronym. Bold lines or vertical spaces between bold lines are the shares in the knowledge source, the thin lines are the closest-in-time Flemish parliament shares of the vote. The numbers below the graphs are the numbers of unique actors "active at the beginning of the year" in that knowledge source..